\documentclass{article}

\usepackage{graphicx} 
\usepackage{subfigure} 

\usepackage{natbib}

\usepackage{hyperref}

\usepackage[accepted]{icml2012}

\makeatletter \renewcommand{\Notice@String}{In \textit{5th International Workshop on Machine Learning and Music}, Edinburgh, Scotland, UK, 2012. Copyright 2012 by the author(s)/owner(s).} \makeatother

\icmltitlerunning{A Mixed Observability Markov Decision Process Model for Musical Pitch}

\begin{document} 

\twocolumn[
\icmltitle{A Mixed Observability Markov Decision Process Model for \\Musical Pitch}

\icmlauthor{Pouyan Rafiei Fard}{rafieifard@ce.sharif.edu}
\icmladdress{Department of Computer Engineering,
            Sharif University of Technology, Tehran, IRAN}
\icmlauthor{Keyvan Yahya}{kxy054@bham.ac.uk}
\icmladdress{School of Psychology, University of Birmingham, Birmingham, UK}

\icmlkeywords{boring formatting information, machine learning, ICML}

\vskip 0.3in
]

\begin{abstract} 
Partially observable Markov decision processes have been widely used to provide models for real-world decision making problems. In this paper, we will provide a method in which a slightly different version of them called Mixed observability Markov decision process, MOMDP,  is going to join with our problem. Basically, we aim at offering a behavioural model for interaction of intelligent agents with musical pitch environment and we will show that how MOMDP can shed some light on building up a decision making model for musical pitch conveniently.\end{abstract} 

\section{Introduction}
\label{introduction}
Partially observable Markov decision processes (POMDPs) have been widely used to provide models for real-world decision making problems. They provide a mathematical framework to model the interaction between the agent and its environment. One of the most notable characteristics of POMDPs is their ability to keep planning in dynamic environments and under uncertainty \cite{Ong10}. To our knowledge, only a few authors have previously mentioned MDPs and POMDPs in the field of computer music. Among them, we could mention \cite{Martin10} who demonstrated the use of POMDPs to control musical behaviour in different conditions.

In this paper, we propose a novel model for interaction of the agents with musical pitch environment based on a variant of POMDPs called mixed observability Markov decision process \cite{Ong10}. First, we mention the theoretical background of our work. In section 3, we propose our model for musical pitch based on MOMDPs. Section 4 addresses some implementation issues and presents an experiment to evaluate our model. Finally, we make our concluding remarks and discuss about the prospective potential developments of this models and its applications.

\section{The Basic Idea of MOMDP} 

Beside the standard models of POMDP, there is a model called MOMDP that makes a slightly different with the former one. The latter is basically a factored POMDP which benefits from factorizing its states. In a MOMDP model, a state \begin{math}s\end{math} is factored into two different variables \begin{math} x,y \end{math}. So by writing  \begin{math} s=(x,y) \end{math} we mean that \begin{math} s \end{math} is consisted of two variables such that \begin{math} x \end{math} stands for fully observable state and \begin{math} y \end{math} stands for partially observable state. Thus having been factorized, we would have a mixed system space \begin{math} S= X \times Y \end{math}where \begin{math} X \end{math} is the state of all values for \begin{math} x \end{math} and either does \begin{math} Y \end{math} for \begin{math} y  \end{math}.
\begin{figure}[ht]
\begin{center}
\centerline{\includegraphics[width=1\columnwidth]{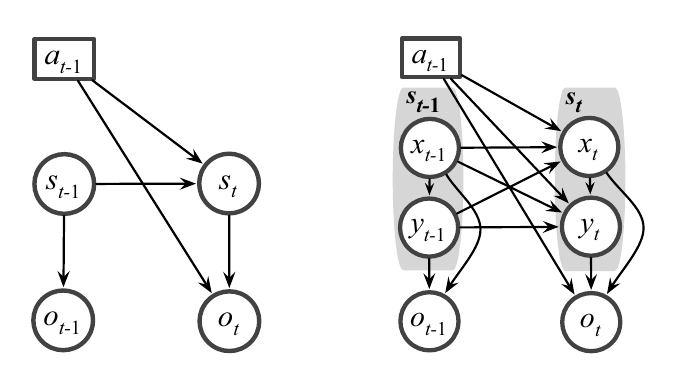}}
\caption{the Standard POMDP model (left) and the MOMDP model (right) in which a state is divided into a fully-observable state \begin{math}x\end{math} and a partially-observable state \begin{math}y\end{math} (adapted from \cite{Ong10}).}
\label{PMOMDP}
\end{center}
\vskip -0.2in
\end{figure} 

\section{The Proposed Model}\label{sec:model}
From music theory, we know that any compound interval can be decomposed into some octaves and a simple interval which this idea can also be brought to any other simple intervals. In our MOMDP-based model, the agent makes its decisions according to the states that it receives from the environment which is here the musical pitch space. A MOMDP is denoted by the tuple \begin{math}(X,Y,A,O,T_x,T_y,Z,R,\gamma)\end{math}. The relationship between these quantities and the musical concepts of our model are given elaborately in the following. 

At each time step the environment is in a state \begin{math}s \in S\end{math} where \begin{math}s=(x,y)\end{math} and \begin{math}x \in X\end{math} is a fully observable state whereas \begin{math}y \in Y\end{math} is a partially observable one.  In our model, a fully observable state represents a musical pitch in which the agent is having a precise estimate of its frequency at the time \begin{math}t \end{math} plus the interval the agent is supposed to make. For the sake of simplicity, we only consider the natural musical pitches and the main  intervals not beyond the octave interval. So, we have \begin{math}S=\{'C','D','E','F','G','A','B'\} \times \{'1st','2nd','3rd',..,'7th'\}.A \end{math} is the set of actions available to the agent. Here an action \begin{math}a \in A\end{math} stands for a making a transition via a musical interval decomposition. In each state we define the possible actions with a set of decompositions. Relatively, the environment lies in partially observable states \begin{math}y \in Y\end{math} as the intermediate state regarding which one of actions the agent makes. Technically, the space of partially observable states is the same space for fully observable states. The parameter \begin{math}O\end{math} is a set of observations that the agent makes which is the possible values of this parameter is the same as values from \begin{math}X\end{math} and \begin{math}Y\end{math}. Finally, The \begin{math}R\end{math} parameter is \begin{math}R_1\gamma +R_2\end{math}, where \begin{math}\gamma \end{math} is the discount factor while \begin{math}R_1\end{math} is the reward for this first interval decomposition and \begin{math}R_2\end{math} is the reward given to the second one.

\vskip -0.2in

\section{Experiment: Reinforcing patterns}\label{sec:experiments}

For testing our model, we developed a Q-Learning algorithm \cite{Watkins89} to perform a similar task to which was done in \cite{Cont08}. We made interactions with the system by feeding a relative pitch pattern as depicted in Figure 2., into the system. For this learning experiment, we set the learning parameters \begin{math}\alpha =0.4, \gamma =0.5\end{math} and \begin{math}N=20 \end{math} as the number of interactions. For a better demonstration of musical learning, the results are presented as intervals and notes. Thus, the y-axis of Figure 3. indicates the intervals and the x-axis is for the notes and for each interval-note pair. The gray-scale values show the learned Q-value and the intensity of these them shows the policy learned by the agent. Also, the values indicated with red rectangles are the values which was originally fed into the system via the pitch contour. 

\begin{figure}[ht]
\vskip -0.1in
\begin{center}
\centerline{\includegraphics[width=0.7\columnwidth,height=16mm]{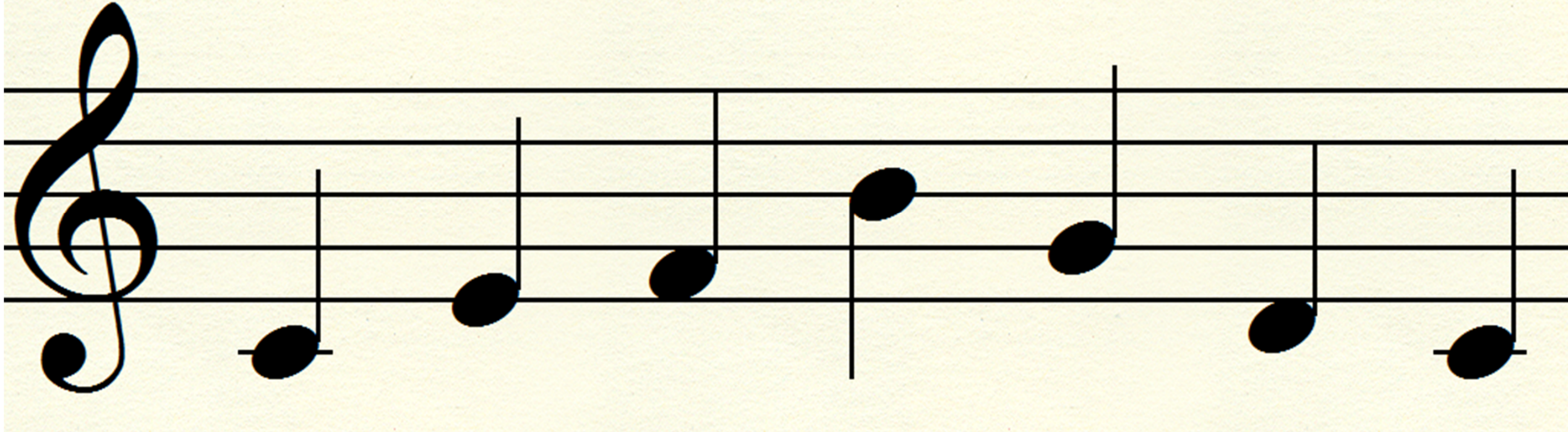}}
\vskip -0.2in
\caption{Pitch contour pattern used in the experiment.}
\label{PitchPattern}
\end{center}
\vskip -0.4in
\end{figure} 

\begin{figure}[ht]
\begin{center}
\centerline{\includegraphics[width=1\columnwidth]{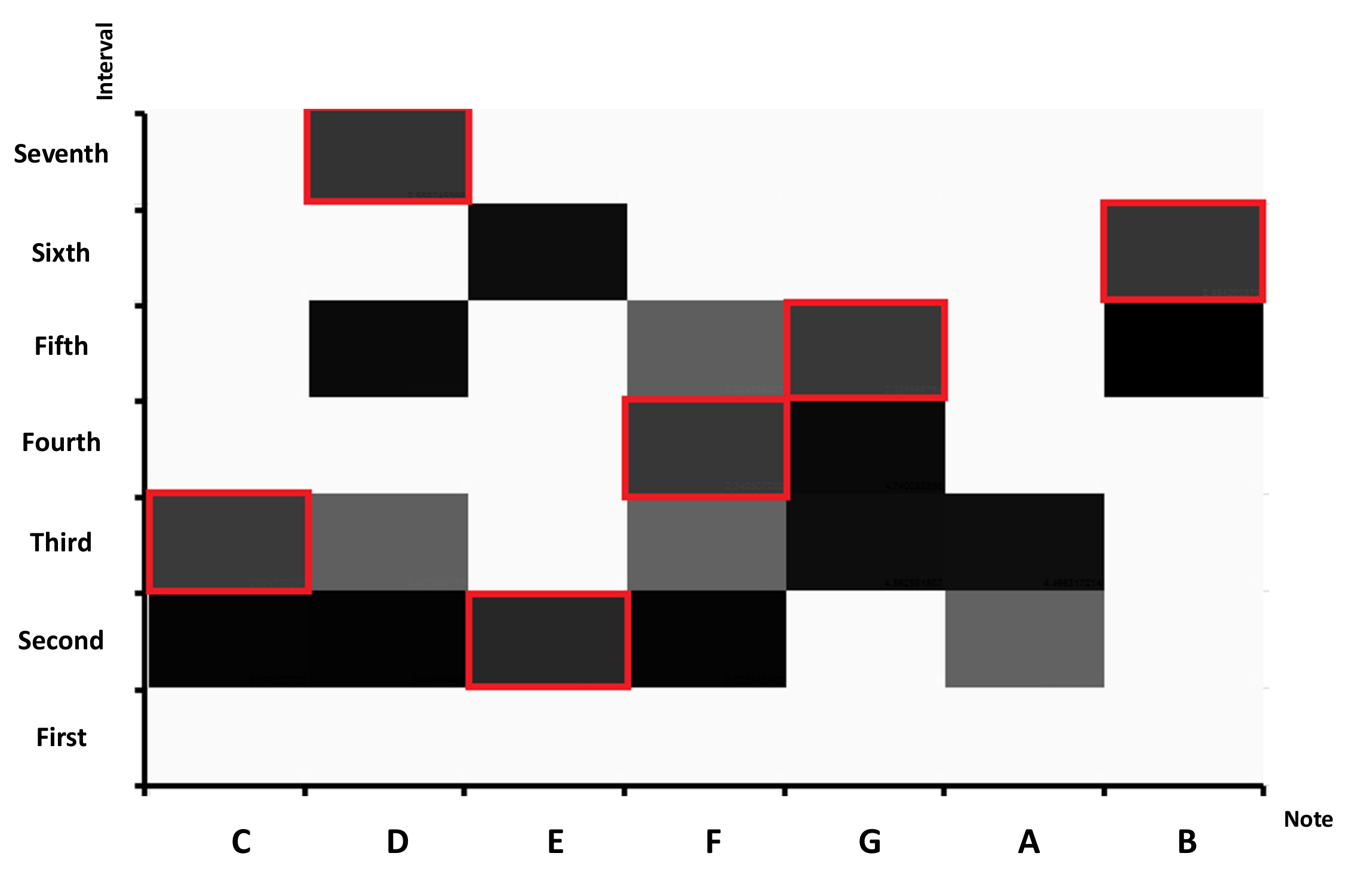}}
\vskip -0.2in
\caption{The results of the experiment.}
\label{Result}
\end{center}
\vskip -0.4in
\end{figure} 

\section{Conclusion}\label{sec:conclusion}

The results of our experiments imply that our agent efficiently learned a behaviour policy. In addition, from Figure 3. we can see that not only our agent learned the given pitch contour (shown by red rectangles) but also some other state-action pairs. This is mainly happened because our method benefits from factorizing each state into a couple of fully-observable and partially-observable states. So, this approach will obviously help to have a faster convergence of the agent which is interacting with musical pitch environment.  

\bibliography{example_paper}

\begin{thebibliography}{8}
\providecommand{\natexlab}[1]{#1}
\providecommand{\url}[1]{\texttt{#1}}
\expandafter\ifx\csname urlstyle\endcsname\relax
  \providecommand{\doi}[1]{doi: #1}\else
  \providecommand{\doi}{doi: \begingroup \urlstyle{rm}\Url}\fi

\bibitem[Ong et~al.(2010)]{Ong10}
Ong, S.~C.~W., Png, ~S.~W., Hsu D., and Lee, W.~S.
\newblock Planning under Uncertainty for Robotic Tasks with Mixed Observability.
\newblock \emph{International Journal of Robotics Research}, 29\penalty0
  (8):\penalty0 1053--1068, 2010.

\bibitem[Martin et~al. (2010)]{Martin10}
Martin, A., Jin, C., van Schaik, A., and Martens, W.~L.
\newblock Partially Observable Markov Decision Processes for Interactive Music Systems.
\newblock In \emph{Proceedings of the International Computer Music
  Conference (ICMC 2010)}, pp.\  490--493, New York, 2010. 

\bibitem[Watkins(1989)]{Watkins89}
Watkins, C.~J.~C.~H.
\newblock \emph{Learning from Delayed Rewards}.
\newblock PhD thesis, Cambridge University, 1989.

\bibitem[Cont(2008)]{Cont08}
Cont, A.
\newblock \emph{Modeling Musical Anticipation: From the time of music to the music of time}.
\newblock PhD thesis, University of Paris 6 and University of California in San Diego, 2008.

\end{thebibliography}
\bibliographystyle{icml2012}

\end{document}